# Multi-stream deep learning framework to predict mild cognitive impairment with Rey Complex Figure Test

Junyoung Park[1,2], Eun Hyun Seo[3,4], Sunjun Kim[5], SangHak Yi[6], Kun Ho Lee[3,7,8*], Sungho Won[2,9,10,11*]

**Affiliations**

[1]Department of Neurology and Neurological Sciences, Stanford University School of Medicine, Stanford, CA, 94305, USA.

[2]RexSoft Inc, Seoul 08826, Korea.

[3]Gwangju Alzheimer's & Related Dementia Cohort Research Center, Chosun University, Gwangju 61452, Korea.

[4]Premedical Science, College of Medicine, Chosun University, Gwangju 61452, Korea

[5]Neurozen Inc., Seoul 06160, Korea.

[6]Department of Neurology, Wonkwang University Hospital, Wonkwang University School of Medicine, Iksan, 15865, Korea.

[7]Department of Biomedical Science, Chosun University, Gwangju 61452, Korea.

[8]Korea Brain Research Institute, Daegu 41068, Korea.

[9]Department of Public Health Sciences, Graduate School of Public Health, Seoul National University, Seoul 08826, Korea.

[10]Interdisciplinary Program in Bioinformatics, Seoul National University, Seoul 08826, Korea.

[11]Institute of Health and Environment, Seoul National University, Seoul 08826, Korea.

**\* Corresponding Authors:**

Sungho Won: won1@snu.ac.kr

Kun Ho Lee: leekho@chosun.ac.kr

# ABSTRACT

Drawing tests like the Rey Complex Figure Test (RCFT) are widely used to assess cognitive functions such as visuospatial skills and memory, making them valuable tools for detecting mild cognitive impairment (MCI). Despite their utility, existing predictive models based on these tests often suffer from limitations like small sample sizes and lack of external validation, which undermine their reliability. We developed a multi-stream deep learning framework that integrates two distinct processing streams: a multi-head self-attention based spatial stream using raw RCFT images and a scoring stream employing a previously developed automated scoring system. Our model was trained on data from 1,740 subjects in the Korean cohort and validated on an external hospital dataset of 222 subjects from Korea. The proposed multi-stream model demonstrated superior performance over baseline models (AUC = 0.872, Accuracy = 0.781) in external validation. The integration of both spatial and scoring streams enables the model to capture intricate visual details from the raw images while also incorporating structured scoring data, which together enhance its ability to detect subtle cognitive impairments. This dual approach not only improves predictive accuracy but also increases the robustness of the model, making it more reliable in diverse clinical settings. Our model has practical implications for clinical settings, where it could serve as a cost-effective tool for early MCI screening.

# INTRODUCTION

Drawing tests have been well-documented for their comprehensive assessment capabilities which include evaluating visuospatial skills, visual memory and executive function, and they are commonly used within the elderly population as a cognitive screening tool for dementia, both in clinical and research fields [1]. Among the most prominent drawing tests are the Pentagon Drawing Test (PDT), the Clock Drawing Test (CDT), and the Rey Complex Figure Test (RCFT). The PDT, for example, requires participants to draw two intersecting pentagons with scoring typically binary (fail or success) [2]. The CDT assesses executive function and visuospatial skills by having subjects draw a clock face set to a specific time, with scoring methods varying significantly – from a binary system to detailed point assignments based on accuracy of contour, number sequence, and hand placement [3-5]. The RCFT, designed by Rey [6], challenges participants to copy and recall a complex figure, with a widely used 36-point scoring system developed by Osterrieth [7].

Recent advancements have seen the application of machine learning approaches to enhance the predictive accuracy of cognitive status from these tests. This is particularly valuable because of the simplicity of administering drawing tests, which could be useful for screening early stages of dementia in clinical fields. For example, deep-learning approaches have been utilized for the digitized PDT[8], CDT [9] and RCFT[10] to predict MCI and CN patients. Additionally, multi-dimensional kinematic parameters extracted from a digital pen and tablet during RCFT were analyzed using logistic regression [11].

However, there are some limitations in previous studies. Primarily, most of these studies had small samples sizes and lacked an external test set, which undermined the reliability of model performances. Even in cases where sample sizes were not small, the performance of models was not sufficiently robust for screening early stages of dementia. This could be attributed to the challenges inherent in utilizing image data in deep learning models. For instance, image data often contains a vast amount of information but can also be prone to noise due to its high dimensionality [12, 13]. Moreover, image data encompasses diverse patterns and features, making it challenging for models to learn effectively, especially when sample sizes are not significantly large [14].

In this paper, we proposed a novel multi-stream deep learning network that combines a spatial stream with raw image data and a scoring stream utilizing an automated scoring system developed in a previous study [15]. The proposed model was implemented by using a total 1,740 subjects (CN 947, MCI 793) to train a deep learning model for distinguishing MCI patients from CN subjects. Additional 222 subjects (CN 106, MCI 116) were utilized as an external dataset to improve the reliability of the model performance.

## MATERIALS AND METHODS

**Datasets**

**GARD cohort**

We enrolled 1,740 subjects from the Gwangju Alzheimer's and Related Dementia (GARD) cohort registry at Chosun University in Gwangju, Korea during 2015-2019. The diagnostic criteria for CN and MCI have been described in Seo et al. [16]. Briefly, CN subjects were included if they were aged 60 or older, had a Clinical Dementia Rating (CDR) score of 0, and exhibited normal cognitive function, with all neuropsychological test z-scores above −1.5 × standard deviation (SD) based on age, education, and gender norms. MCI patients were aged 60 or older, had a CDR score of 0.5, and met the MCI criteria established by [17].

**WUH cohort**

The Wonkwang University Hospital (WUH) cohort includes 106 CN subjects and 116 MCI patients enrolled between 2017 and 2022. In alignment with our training set criteria, subjects were classified based on their CDR scores: a CDR score of 0 indicated a CN diagnosis, while a score of 0.5 indicated MCI.

**Deep learning architecture**

Figure 1A provides an overview of the proposed method. Our model predicts the probability of an individual being classified as a MCI patients using three pre-processed RCFT images along with age, sex and years of education. The pre-processing method for the RCFT images follows the protocol outlined by Park et al. [15]. Our prediction model employs a dual-stream architecture: a spatial stream and a scoring stream. Both streams process data through softmax functions, and their outputs are merged using average fusion to yield the final classification probability. In the spatial stream, each 512x512 image is input into a CNN model that uses EfficientNet [18] as its backbone. We selected EfficientNet-B2 for its efficiency and suitability in medical applications, given its lower parameter count and adequate performance with limited datasets. EfficinetNet-B2 incorporates a 3x3 convolution layer followed by multiple 3x3 and 5x5 mobile inverted bottleneck convolution (MBConv) blocks, a design borrowed from MobileNet [19] (Figure 1B). Post-CNN, the feature map are flattened, and a multi-head self-attention layer is applied, enhancing the model's focus on significant spatial region. The multi-head self-attention mechanism, as defined by [20], combines multiple self-attention layers to capture diverse features, expressed as:

$$MultiHead(Q, K, V) = Concat(head_1, \ldots, head_h)W^O,$$

$$head_i = Attention(QW_i^Q,\ KW_i^K, VW_i^V)$$

where $Q, K, V$ are the query, key and value matrix, respectively, and we use four attention heads ($h$=4). The outputs from multi-head self-attention layers are integrated and processed through two fully connected (FC) layers followed by a softmax function.

Conversely, the scoring stream uses a previously developed deep learning model [15] to predict RCFT scores. The scores for three images, along with demographic data, are concatenated and passed through an FC layer with a softmax function. It is important to note that the scoring model's weights remain fixed during training, preventing updates.

[Figure 1]

**Baseline models**

The proposed model was evaluated against four baseline models: three logistic regression models and one deep learning model. The first baseline model utilized MMSE scores. The second and third models used three RCFT scores, scored by trained experts and a previous AI scoring system, respectively. The final baseline was a deep learning model, which solely utilized the spatial stream network. All baseline models included age, sex and years of education as covariates.

**Scoring validation**

To mitigate human errors in scoring, scanning and digitizing, we tailored our AI scoring system specifically for the external test set to enhance data quality. For images where the difference between the human expert scores and AI-generated scores exceeded ten points, we conducted a re-examination by trained human experts. Following this, we compared the AI-generated scores with these newly corrected scores to ensure accuracy and reliability.

**Experiments**

We conducted prediction model building and performance evaluation using data from GARD and WUH cohort. GARD cohort was employed to construct the prediction model. Throughout the training process, we utilized the binary cross-entropy as the loss function and the Adam optimizer was adopted to minimize the loss function. To prevent overfitting, we reduced the initial learning rate to 10% every five epochs and implemented early stopping if there was no improvement in validation loss after 30 epochs, ensuring that the final model weights selected corresponded to the lowest validation loss.

To evaluate our model's performance, GARD cohort was randomly divided into training, validation and test sets with 6:2:2 ratio. This division process was repeated fifty times. External validation was performed using WUH cohort. Model performance was assessed using the area under receiver operating characteristics (AUC), the accuracy (ACC), sensitivity (SEN) and specificity (SPE).

All experiments were conducted using the Pytorch library (v 2.0.0) in Python (v 3.8.8) with NVIDIA 1080ti GPUs with 48 GB of memory per GPU.

## RESULTS

**Characteristics**

Table 1 summarizes the clinical characteristics of subjects in the GARD and WUH cohort datasets. In the GARD dataset, the average ages were 71.8 ($\pm$6.1) years for CN subjects and 73.5 ($\pm$6.4) years for MCI patients (P<0.01). Education levels and MMSE scores also significantly differed between CN subjects (education level: 10.4$\pm$4.6; MMSE score: 27.5$\pm$2.1) and MCI patients (9.8$\pm$4.7; 25.5$\pm$3.1) (P<0.01). Similarly, sex ratios exhibited comparable trends in both groups. Conversely, the WUH dataset revealed no significant differences in the average ages between CN (69.9±7.7) subjects and MCI (71.4±8.3) patients (P>0.05), nor were there differences in education levels between CN (8.7±4.2) and MCI (9.2±4.5) groups (P>0.05). Comparing the two datasets, the external test set consistently showed lower age, education level, and RCFT scores across both groups, with the exception of the education level and RCFT copy score in CN group of the GARD dataset.

**Scoring validation**

The initial correlation ($R^2$) between scores by AI and those by experts was 0.81, with a mean absolute error (MAE) of 3.0 point (Figure 2A). Discrepancies exceeding 10 points between the ground truths and predicted scores were identified in 30 images. Upon validation, scores for 26 of these images were corrected. After these adjustments, the correlation improved significantly to an $R^2$ of 0.95 with an MAE = 2.0 (Figure 2 B).

[Figure 2]

**Comparison of model performance via internal test using GARD cohort**

We evaluated the classification performances of five models, including three that incorporated the proposed method. These models are: 1) logistic regression using MMSE scores; 2) logistic regression using RCFT scores assessed by experts; 3) logistic regression using RCFT scores predicted by the AI model; 4) deep learning model utilizing only spatial stream network; 5) deep learning model employing multi stream networks. The mean performances of those models are shown in Table 2 (A).

The logistic regression model with MMSE scores demonstrated the lowest performance, with an AUC of 0.714 [95% confidence interval: 0.706-0.712], an ACC of 0.660 [0.652-0.667], SEN of 0.625 [0.613-0.636] and SPE of 0.694 [0.685-0.704]. The logistic regression model using expert-assessed RCFT scores recorded an AUC of 0.776 [0.768-0.782], an ACC of 0.705 [0.699-0.712], an SEN of 0.700 [0.689-0.711] and an SPE of 0.71 [0.700-0.722]; the performance of the model using AI-predicted RCFT scores was similar, with an AUC of 0.777 [0.770-0.783], ACC of 0.710 [0.703-0.717], SEN of 0.699[0.689-0.709] and SPE of 0.721 [0.710-0.731].

Performance improvements were evident with the spatial stream network model, which achieved an AUC of 0.803 [0.768-0.837], ACC of 0.731 [0.702-0.761], SEN of 0.701 [0.661-0.741] and SPE of 0.762[0.720-0.804]. Finally, our proposed deep learning model using the two-stream network outperformed all baseline models across all metrics, with an AUC of 0.852 [0.837-0.869], ACC of 0.771 [0.755-0.787], SEN of 0.742 [0.718-0.767] and SPE of 0.800 [0.774-0.823].

**External validation using WUH cohort**

Performance metrics for the trained models on this set are detailed in Table 2 (B). The logistic regression model using expert-rated RCFT scores from the initial dataset demonstrated an AUC of 0.750 [0.750-0.751], ACC of 0.709 [0.707-0.712], SEN of 0.832 [0.829-0.835] and SPE of 0.575 [0.571-0.579]. With the validated dataset based on the re-rated RCFT scores, the model's performance improved to an AUC of 0.813 [0.812-0.814], ACC of 0.750 [0.748-0.753], SEN of 0.799 [0.718-0.767] and SPE of 0.800 [0.774-0.823]. The logistic model with AI-predicted RCFT scores displayed comparable performance to that of human experts (AUC=0.804[0.803-0.805], ACC=0.722[0.721-0.725], SEN=0.799[0.797-0.802] and SPE=0.639[0.634-0.722]).

The deep learning model employing the spatial stream network achieved a higher AUC (0.837[0.814-0.860]), ACC (0.744[0.719-0.768]) and SPE (0.745[0.697-0.792]) but had a lower SEN (0.743[0.690-0.800]). Our proposed deep learning method using the two-stream network outperformed all baseline models, showing superior performance across all metrics: AUC=0.872[0.862-0.882], ACC=0.781[0.768-0.795], SEN=0.836[0.807-0.864] and SPE=0.722[0.687-0.757].

**[Figure 3]**

## DISCUSSION

In this article, we developed a multi-stream deep learning network to differentiate between MCI patients and CN subjects. Our approach surpasses previous methods utilizing drawing test (PDT, CDT and RCFT) by leveraging a larger sample size and an external test set, thereby enhancing the robustness and performance of the model. Notably, our model outperformed existing studies, achieving the highest recorded performance metrics.

Our multi-stream network combines both the scoring stream and spatial stream. The scoring stream incorporates an AI scoring system for RCFT, which save time and human resources while proactively preventing human errors, thus improving accuracy. This improvement was evidenced by results showing that the model, when AI scoring was used for QC, exhibited much higher performance compared to the model performance using the initial expert-assessed RCFT scores without QC. Furthermore, while it takes approximately 5 minutes for an expert to score one subject, our AI scoring system takes only 10 seconds. The spatial stream of our model utilizes raw RCFT images as input, and extracts subtle details within the images, such as pen thickness and shape, which are not captured by the human scoring system (ranging from 0-36 points). This leads to substantial improvement in performance compared to models that rely solely on scoring. However, although raw image data is rich with information, it also includes considerable noise; therefore, the integration of multi-head self-attention layers helps the model to prioritize crucial spatial regions within the feature map, boosting performance. However, models that rely solely on raw images have shown higher SDs in performance compared to logistic models utilizing scores, and the performance of the spatial stream network may be compromised due to differences in resolution between existing training images and new test images. By combining the advantages of both scoring stream network, which utilizes human scoring systems, and the spatial stream network, which processes images, our proposed method achieves high and robust performance.

The proposed method offers a cost-effective and efficient screening tool for MCI patients at the medical check-up centers. Currently, the MMSE is the most popularly utilized screening tool, known for its simplicity and quick administration time of approximately 5-10 minutes [2]. However, our results indicate that MMSE is less informative for predicting MCI and lacked accuracy in distinguishing between CN subjects and MCI patients (AUC = 0.714). Another study reported MMSE performance with an AUC of 0.733 (N=2,577) [8]. In contrast, comprehensive cognitive function tests such as Neuropsychological Test Battery are more time-consuming, taking up 2 hours to administer [21] and pose challenges in examining multiple subjects due to the additional time required for scoring and interpretation. Although the RCFT requires more times than the MMSE, approximately 30 minutes including a 20-minute delay interval [22], our model based on the RCFT significantly outperformed that of the MMSE (AUC>0.85). Furthermore, since our model does not necessitate additional time for expert scoring, it is highly efficient compared to other cognitive function tests that rely on expert scoring.

Despite the flexibility of the proposed method, our study had some limitations and areas for future development. First, we did not incorporate additional ancillary information beyond the raw images. Recent studies have shown that kinematic data such as pressure, velocity, time which cannot be captured by traditional paper-and-pencil drawing tests but recorded by tabled-based tests revealed significant differences between case and control groups. These parameters suggest potentially useful covariates to enhance the performance of prediction models [11, 23]. We have developed a tablet-based application that administers the RCFT, records the drawing process and extracts kinematic parameters. By incorporating this information, further improvement may be possible. Second, verbal tests have also played a crucial role in neuropsychological evaluation [24]. Recent advancements in automatic speech recognition technology, such as BERT [25], have enabled the exploration of speech-based methods for AD detection [26, 27]. For future work, we plan to develop tablet-based, fully automated memory tests that integrate both visual and verbal assessments.

In conclusion, our multi-stream deep learning network outperformed previous studies in distinguishing MCI patients from CN subjects. By integrating human scoring systems and image-based information, our model demonstrated robust performance across internal and external datasets. Our findings suggest potential clinical utility as a time-efficient screening tool for cognitive impairment.

## ACKNOWLEDGEMENT

Not applicable

## FUNDING

This work was supported by the Technology Innovation Program (20022810, Development and Demonstration of a Digital System for the evalution of geriatric Cognitive impairment) funded By the Ministry of Trade, Industry & Energy (MOTIE, Korea), and by the "Korea National Institute of Health"(KNIH) research project No.#2024ER210800

## AVAILABILITY OF DATA AND MATERIALS

The dataset for the current study is not publicly available but is available from the corresponding author upon reasonable request.

## ETHICS APPROVAL

The study was approved by the Institutional Review Boards of Chonnam National University Hospital (CNUH-2019-279) and Wonkwang University Hospital (2022-01-024-004). Written informed consent was obtained from each participant or their legal guardian.

## CONSENT FOR PUBLICATION

Not applicable.

## COMPETING INTERESTS

The authors declare that they have no competing interests.

# Figures and Tables

**Figure 1. Model Architecture.** (A) Overall model architecture featuring a dual-stream design: a spatial stream with EfficientNet-B2 and a scoring stream using a pre-trained model for RCFT scoring. The combined outputs undergo average fusion to produce the final CN or MCI classification. (B) Detailed architecture of the EfficientNet-B2 model used in the spatial stream, including convolutional layers and Mobile Inverted Bottleneck Convolution (MBConv) layers.

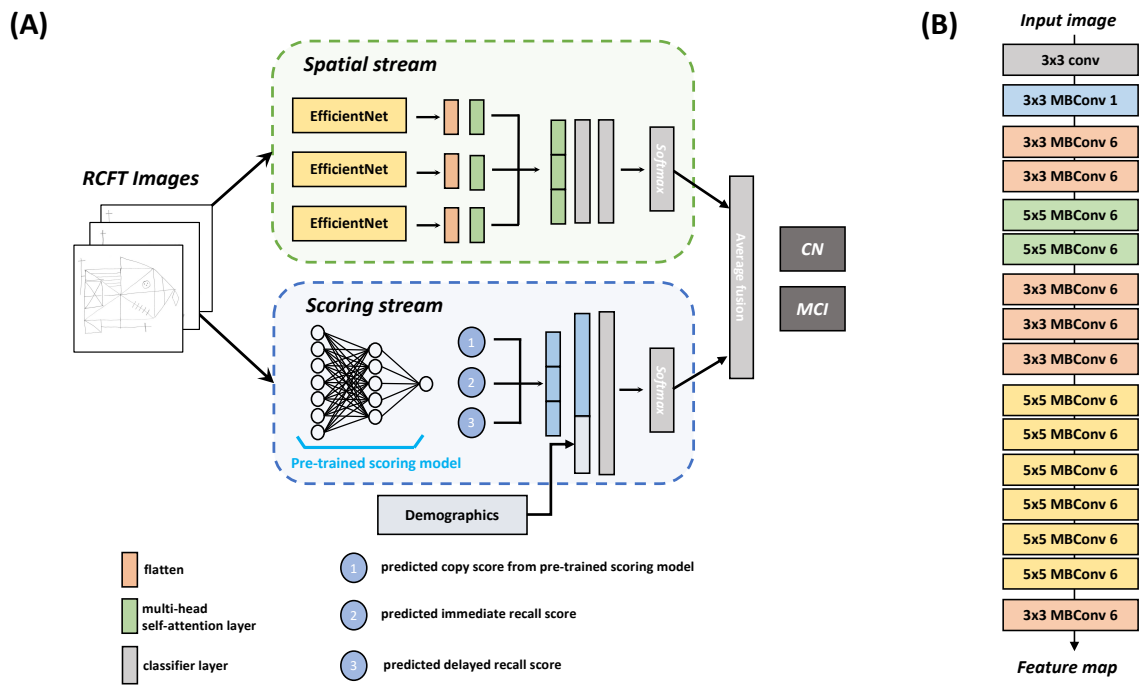

**Figure 2. Comparative validation of AI-assessed and expert-assessed scores.** (A) Results before quality control (QC), where AI-generated scores from a pre-trained model for RCFT scoring were compared to human expert scores. Significant discrepancies (greater than ten points) between the AI-generated scores and human expert scores (highlighted in red) led to re-examination by trained experts. (B) Results after QC, showing improved $R^2$ between AI-generated and expert-corrected scores following the re-examination process.

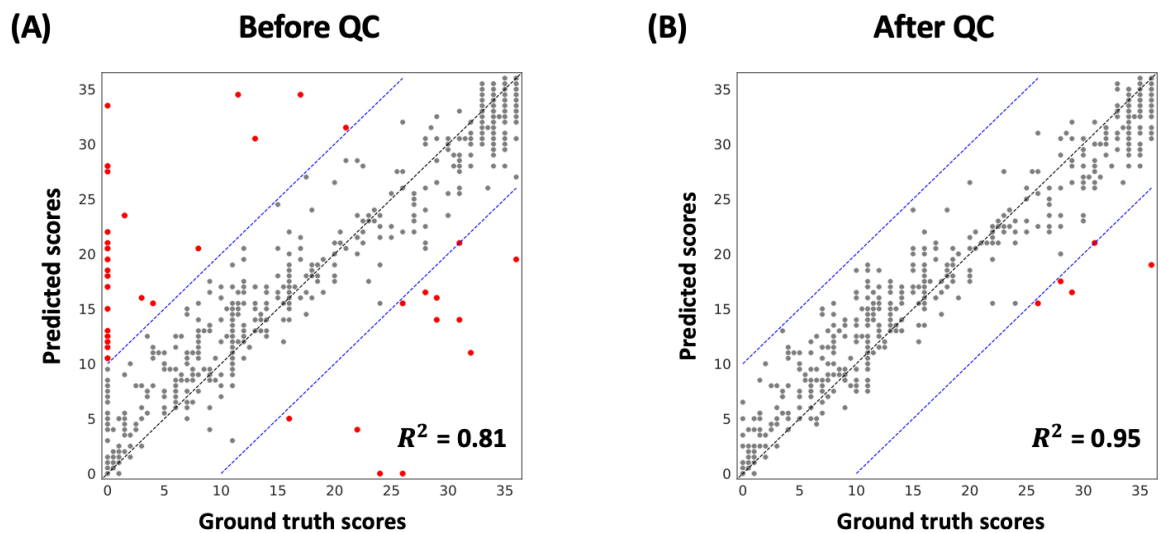

**Figure 3. ROC curve for external test set (WUH cohort dataset).** The ROC curve is plotted using the median AUC results from 50 bootstrap samples, illustrating the performance of different models.

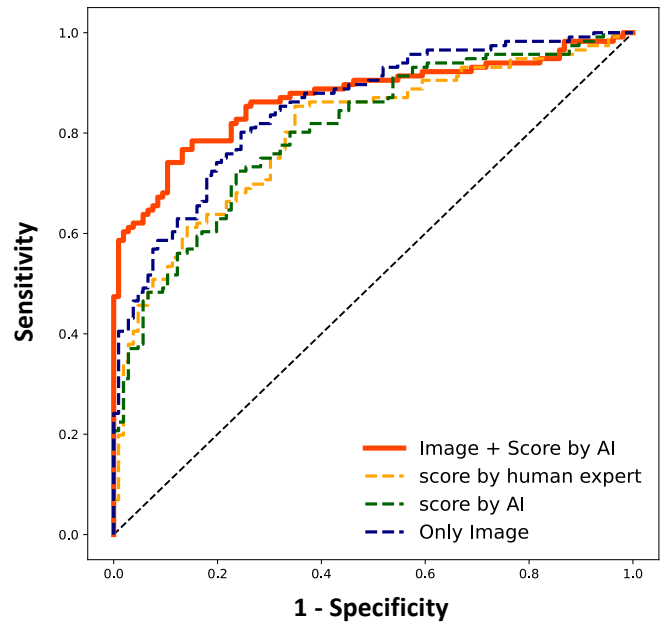

**Table 1. Descriptive statistics.** A dataset of 1,740 subjects from the Gwangju Alzheimer's and Related Dementia (GARD) cohort was used for training, and an external test set of 222 subjects from Wonkwang University Hospital (WUH) was used for validation.

| Dataset | | GARD (training) | | | WUH (test) | | |
|---|---|---|---|---|---|---|---|
| | | **Total (N=1,740)** | **CN (N=947)** | **MCI (N=793)** | **Total (N=222)** | **CN (N=106)** | **MCI (N=116)** |
| Age (years) | | 72.6 (6.3) | 71.8 (6.1) | 73.5 (6.4) | 70.7 (8.0) | 69.9 (7.7) | 71.4 (8.3) |
| Sex N (female, %) | | 743 (57.3%) | 378 (60.1%) | 365 (54.0%) | 140 (63.0%) | 76 (71.7%) | 64 (55.2%) |
| Education (years) | | 10.1 (4.6) | 10.4 (4.5) | 9.8 (4.7) | 9.0 (4.4) | 8.7 (4.2) | 9.2 (4.5) |
| MMSE scores | | 26.5 (2.8) | 27.5 (2.1) | 25.5 (3.1) | - | - | - |
| RCFT Score | copy | 32.0 (5.1) | 33.6 (3.0) | 30.2 (6.4) | 30.6 (9.1) | 33.7 (4.2) | 27.8 (11.2) |
| | immediate | 12.8 (7.4) | 15.7 (6.7) | 9.3 (6.6) | 9.2 (8.5) | 12.7 (8.2) | 6.1 (7.4) |
| | delayed | 12.7 (7.2) | 15.7 (6.3) | 9.1 (6.7) | 8.5 (8.5) | 11.9 (8.1) | 5.3 (7.5) |

**Table 2. Results of model prediction performance**. The baseline models consisted of three logistic regression models using MMSE scores, RCFT scores by experts, and RCFT scores by a previous AI model, respectively, and one deep learning model that utilized only the spatial stream network. All baseline models included chronological age, sex, and education as covariates. The data was split into 6:2:2 (training, validation, and testing sets), and this process was repeated 50 times.

**(A) Internal test using the GARD cohort dataset.**

| Input modality | GARD | | | |
|---|---|---|---|---|
| | AUC | Accuracy | Sensitivity | Specificity |
| MMSE scores | 0.714 [0.706-0.721] | 0.660 [0.652-0.667] | 0.625 [0.613-0.636] | 0.694 [0.685-0.704] |
| RCFT scores by experts | 0.776 [0.768-0.782] | 0.705 [0.699-0.712] | 0.700 [0.689-0.711] | 0.711 [0.700-0.722] |
| RCFT scores by AI | 0.777 [0.770-0.783] | 0.710 [0.703-0.717] | 0.699 [0.689-0.709] | 0.721 [0.710-0.731] |
| Only RCFT images | 0.803 [0.768-0.837] | 0.731 [0.702-0.761] | 0.701 [0.661-0.741] | 0.762 [0.720-0.804] |
| **Image + score by AI (Our method)** | **0.852 [0.837-0.869]** | **0.771 [0.755-0.787]** | **0.742 [0.718-0.767]** | **0.800 [0.774-0.823]** |

**(B) External test using the WUH cohort dataset.** RCFT scores by experts[a] refers to the model using scores from the initial dataset before QC, while RCFT scores by experts[b] indicates the model with the validated dataset after QC based on the re-rated RCFT scores using a previous AI model.

| Input modality | WUH | | | |
|---|---|---|---|---|
| | AUC | Accuracy | Sensitivity | Specificity |
| RCFT scores by experts[a] | 0.750 [0.750-0.751] | 0.709 [0.707-0.712] | 0.832 [0.829-0.835] | 0.575 [0.571-0.579] |
| RCFT scores by experts[b] | 0.813 [0.812-0.814] | 0.750 [0.748-0.753] | 0.849 [0.845-0.852] | 0.643 [0.639-0.648] |
| RCFT scores by AI | 0.804 [0.803-0.805] | 0.722 [0.721-0.725] | 0.799 [0.797-0.802] | 0.639 [0.634-0.644] |
| Only RCFT images | 0.837 [0.814-0.860] | 0.744 [0.719-0.768] | 0.743 [0.690-0.800] | 0.745 [0.697-0.792] |
| **Image + score by AI (Our method)** | **0.872 [0.862-0.882]** | **0.781 [0.768-0.795]** | **0.836 [0.807-0.864]** | **0.722 [0.687-0.757]** |